\documentclass[letterpaper, 10 pt, conference]{ieeeconf}  

\IEEEoverridecommandlockouts                              

\usepackage{enumerate}
\usepackage{color}
\usepackage{xspace}
\usepackage{makecell}

\usepackage[free-standing-units]{siunitx}

\usepackage{float}
\usepackage{svg}
\usepackage{caption}
\usepackage{multicol}
\usepackage{multirow}

\usepackage[utf8]{inputenc} 
\usepackage{url}            
\usepackage{amsfonts}       
\usepackage{nicefrac}       
\usepackage{microtype}      
\usepackage{amsmath}

\usepackage{graphicx}
\usepackage{amssymb}
\usepackage{cancel}
\usepackage{pifont}
\usepackage{subcaption}
\usepackage{wrapfig}
\usepackage[export]{adjustbox}
\usepackage{bbm}
\usepackage{microtype}
\usepackage{graphicx}
\usepackage{booktabs} 
\usepackage[free-standing-units]{siunitx}

\providecommand{\mypara}[1]{{\noindent{\bf #1}}}
\providecommand{\mypara}[1]{{\noindent{\bf #1}}}

\newcommand{\customfootnotetext}[2]{{
  \renewcommand{\thefootnote}{#1}
  \footnotetext[0]{#2}}}

\newcommand{\methodNameLong}{Harmonic Mobile Manipulation\xspace}
\newcommand{\methodName}{\textsc{HarmonicMM}\xspace}
\newcommand{\suite}{Daily Mobile Manipulation Task Suite\xspace}

\title{\LARGE \bf
Harmonic Mobile Manipulation
}

\def\maketitlesupplementary{
{
    \centering
    \Large
    \textbf{Appendix}\\
    \vspace{-0.1in}
}
}

\author{
Ruihan Yang\textsuperscript{1*}\hspace{2mm}
Yejin Kim\textsuperscript{2} \hspace{2mm}
Rose Hendrix\textsuperscript{2} \hspace{2mm}
Aniruddha Kembhavi\textsuperscript{2,3} \hspace{2mm}
Xiaolong Wang\textsuperscript{1} \hspace{2mm} 
Kiana Ehsani\textsuperscript{2} \hspace{2mm}\\
\vspace{1mm}
\textsuperscript{1}UC San Diego\;
\textsuperscript{2}PRIOR @ Allen Institute for AI\;
\textsuperscript{3}University of Washington, Seattle
}

\begin{document}

\twocolumn[{%
\renewcommand\twocolumn[1][]{#1}%
\maketitle
\begin{center}
    \centering 
    \vspace{-0.175in}
    \includegraphics[width=0.95\textwidth]{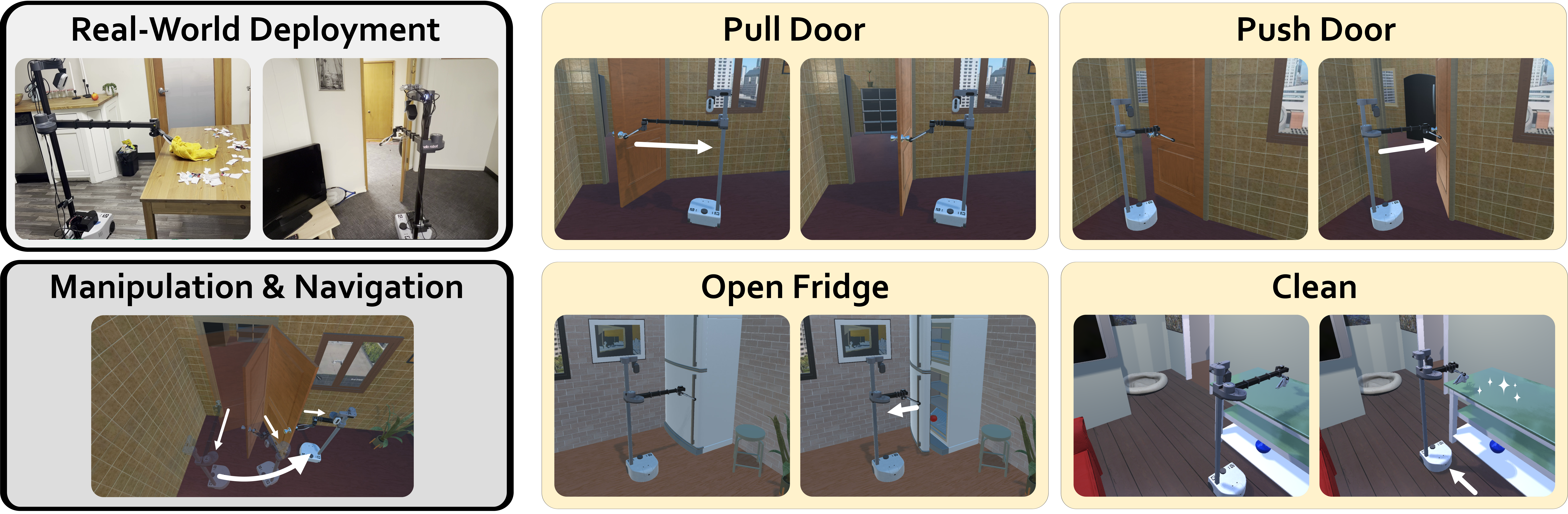}
    \vspace{-0.05in}
    \captionof{figure}{\textbf{Harmonic Mobile Manipulation:} In this work, we address diverse mobile manipulation tasks integral to human's daily life. Trained in a photo-realistic simulation,  Our controller effectively accomplishes tasks through harmonious mobile manipulation in a \textit{real-world apartment} featuring a novel layout, \textit{without any fine-tuning or adaptation}. 
    }
    \label{fig:teaser}
\end{center}%
}]

\customfootnotetext{*}{Work done during internship at PRIOR @ Allen Institute for AI}
\customfootnotetext{}{More results on \url{https://rchalyang.github.io/HarmonicMM}}

\thispagestyle{empty}
\pagestyle{empty}


\begin{abstract}

Recent advancements in robotics have enabled robots to navigate complex scenes or manipulate diverse objects independently.
However, robots are still impotent in many household tasks requiring coordinated behaviors such as opening doors. The factorization of navigation and manipulation, while effective for some tasks, fails in scenarios requiring coordinated actions. 
To address this challenge, we 
introduce, \methodName, an end-to-end learning method that optimizes both navigation and manipulation, showing notable improvement over existing techniques in everyday tasks. This approach is validated in simulated and real-world environments and adapts to novel unseen settings without additional tuning. Our contributions include a new benchmark for mobile manipulation and the successful deployment with only RGB visual observation in a real unseen apartment, demonstrating the potential for practical indoor robot deployment in daily life.

\end{abstract}    
\section{Introduction}
\label{sec:intro}

The field of robot learning has traditionally treated robot navigation and manipulation as separate domains, leading to significant advancements in each. On one hand, agents have been trained to efficiently explore and navigate environments~(\cite{khandelwal2022:embodied-clip, deitke2023phone2proc, gervet2022navigating, yang2023neural}), while on the other, substantial progress has been made in performing complex manipulation tasks, such as handling articulated objects with static arms in tabletop settings~(\cite{mendonca2023alan, wang2023mimicplay, zhu2023learning, dexpoint}).

However, in household settings, many tasks require coordinated body movements while simultaneously manipulating objects with our arms. To address this challenge, recent efforts have focused on enhancing robotic agents with the ability for mobile manipulation. The most prominent approach is invoking navigation and manipulation as separate modules. This separation is achieved either by utilizing a high-level semantic controller, such as a large language model ~(\cite{saycan2022arxiv, huang2022inner}), or by estimating low-level subgoals using motion planners~\cite{xia2021relmogen}. In these works, behaviors are split into distinct navigation and manipulation skills.

While this approach has proven effective for long-horizon pick-and-place tasks, it presents certain limitations. Firstly, it simply fails for tasks that demand simultaneous navigation and manipulation, such as opening a door. Without coordinated action, the robot's body may collide with the door, preventing the successful opening of the door.
Secondly, the physical constraints of the robot can negatively impact task completion. 
For example, cleaning a large table with a robot that has limited movement range, like the Stretch robot with a 3-DOF arm, requires concurrent control of the arm and base. Thirdly, the disjoint approach is highly inefficient. Consider the act of opening a door: when we open a door, we instinctively reach for the handle while simultaneously moving towards the door. Separating arm and leg movements can have a negative impact on efficiency and performance.

To address these challenges, we introduce an approach that efficiently coordinates navigation and manipulation for complex mobile manipulation tasks. Leveraging the effectiveness of training models in procedurally generated simulated environments, as shown by recent studies~\cite{deitke2023phone2proc}, we train our models in the visually diverse environments of ProcTHOR~\cite{deitke2022procthor}. Recognizing the limitations of ProcTHOR, which lacks features like door opening or table cleaning, we have expanded the simulation to include these functionalities. Our end-to-end model, which relies solely on RGB and arm proprioception, demonstrates an absolute improvement of 17.6\% across four tasks compared to existing baselines and successfully transfers to real-world applications.

Our primary contributions are:

\begin{itemize}
\item An end-to-end learning approach that jointly optimizes navigation and manipulation, achieving an absolute improvement of 17.6\% in average success rate across tasks compared to previous methods.
\item Adding the support for more complex tasks, such as door opening and table cleaning, to ProcTHOR.
\item Successful transfer of agents trained in simulation to real-world with only RGB visual observation.
\item Introducing a new benchmark for complex mobile manipulation tasks, including opening fridges, cleaning tables, and opening doors by pulling and pushing.
\end{itemize}

\section{Related Work}
\label{sec:related}

\mypara{Embodied AI Benchmark.} Over the past few years, a variety of standard benchmarks have emerged to assess progress in the field of embodied AI.

These benchmarks primarily focus on high-level planning, scene understanding, and basic interaction tasks like object navigation and house rearrangement~(\cite{ai2thor,deitke2022procthor, habitatchallenge2023, yadav2022habitat, habitat19iccv, yenamandra2023homerobot, puig2023habitat, ehsani2021manipulathor, gan2021threedworld}) and generally limit physical world interactions to straightforward pick-and-place actions. 
Deitke et al.\cite{deitke2022procthor} have created diverse houses for training embodied agents to navigate through various everyday scenes. 
Li et al.\cite{li2022behaviork} tackles various household tasks using predefined motion primitives. 
In contrast to these benchmarks, our work extends beyond simple pick-and-place tasks. Our robot is designed to perform complex mobile manipulations, requiring tight coordination between navigation and manipulation. 

\mypara{Robotics Manipulation.}
In robotic manipulation, the focus has traditionally been on either fixed-base tabletop tasks or elementary mobile manipulations in controlled environments~(\cite{yu2019meta, gu2023maniskill2, mu2021maniskill, brohan2023rt1, brohan2023rt2, generalpatternmachines2023, urakami2022doorgym, bahl2022human, pari2021surprising}). Mu et al.\cite{mu2021maniskill} have developed benchmarks for straightforward mobile manipulation tasks in open spaces. Gu et al.\cite{gu2023rttrajectory} have demonstrated how robots can clean tables using imitation learning in a tabletop setup. While some works have explored basic mobile manipulations, such as door opening in lab settings \cite{urakami2022doorgym}, our efforts extend to more dynamic tasks requiring the robot to actively navigate and interact with its environment, pushing the boundaries of traditional mobile manipulation research.

\mypara{Mobile Manipulator.}
In the realm of mobile manipulation, previous studies have designed task-specific mobile manipulators using optimization techniques\cite{lew2022robotic} or employing Task-and-motion-planning (TAMP) methods (\cite{6906922, garrett2020pddlstream}). Lew et al.\cite{lew2022robotic} developed a table-cleaning controller using whole-body trajectory optimization. While effective for their intended purposes, these methods tend to be narrowly focused with limited scalability and adaptability to different scenarios. Our pipeline could apply to a wide range of tasks, beyond the constraints of task-specific methodologies and offer a more flexible solution for various mobile manipulation challenges.

Another significant line of work involves learning-based methods (\cite{xia2021relmogen, gu2023multiskill}), which offer better generalization in task completion. These methods diverge into two sub-categories: short-horizon whole-body control of robots (\cite{sun2021fully, hu2023causal, deep-wbc}), and long-horizon tasks solved at the scene level with predefined motion primitives in an iterative or two-stage manner (\cite{yokoyama2023asc, xia2021relmogen, li2019hrl4in, Jauhri_2022, saycan2022arxiv, brohan2023rt1, brohan2023rt2}). 
Yokoyama et al.\cite{yokoyama2023asc}, Xia et al.\cite{xia2021relmogen} and Ahn et al.\cite{saycan2022arxiv} exemplify solving long-horizon mobile manipulation tasks using predefined navigation and manipulation skills, either iteratively or in a two-stage approach. 
Notably, our work proposes a unified learning framework that integrates navigation and manipulation in a seamless manner, addressing the limitations of predefined primitives by focusing on learning adaptable skills for complex, everyday tasks.

\section{\suite}
\label{sec:tasks}

\begin{figure*}[t]
    \includegraphics[width=0.6\textwidth,clip]{
        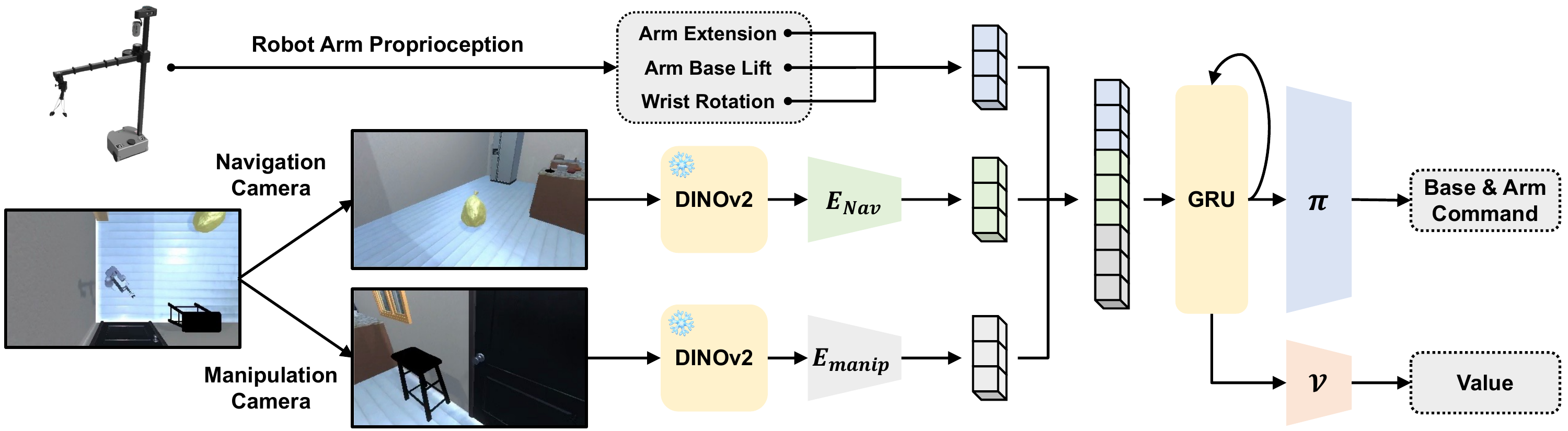
    }
    \hfill
    \includegraphics[width=0.38\textwidth]{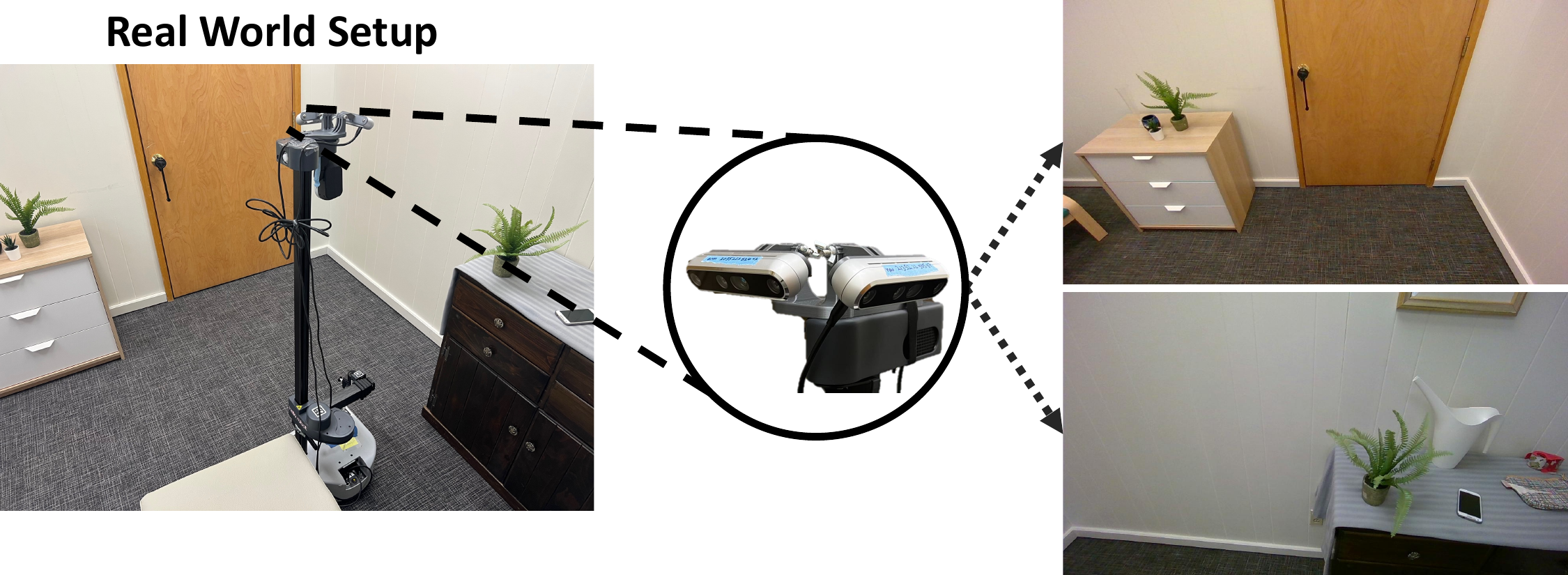}
    \vspace{-0.05in}
    \caption{\small{\textbf{\methodName Network Architecture (Left):} Our \methodName controller takes robot proprioception and multi-view visual observations as input and output navigation and manipulation commands at the same time. 
    \textbf{Real Visual Observations (Right):} Our robot is shown on the Left and the observations from \textit{Nav Cam} and \textit{Manip Cam} are shown on the Top Right and Bottom Right respectively.
    }}
    \label{figure:method}
\vspace{-0.2in}
\end{figure*}

\begin{figure}[t]
    \begin{center}
        \includegraphics[width=0.4\textwidth,clip]{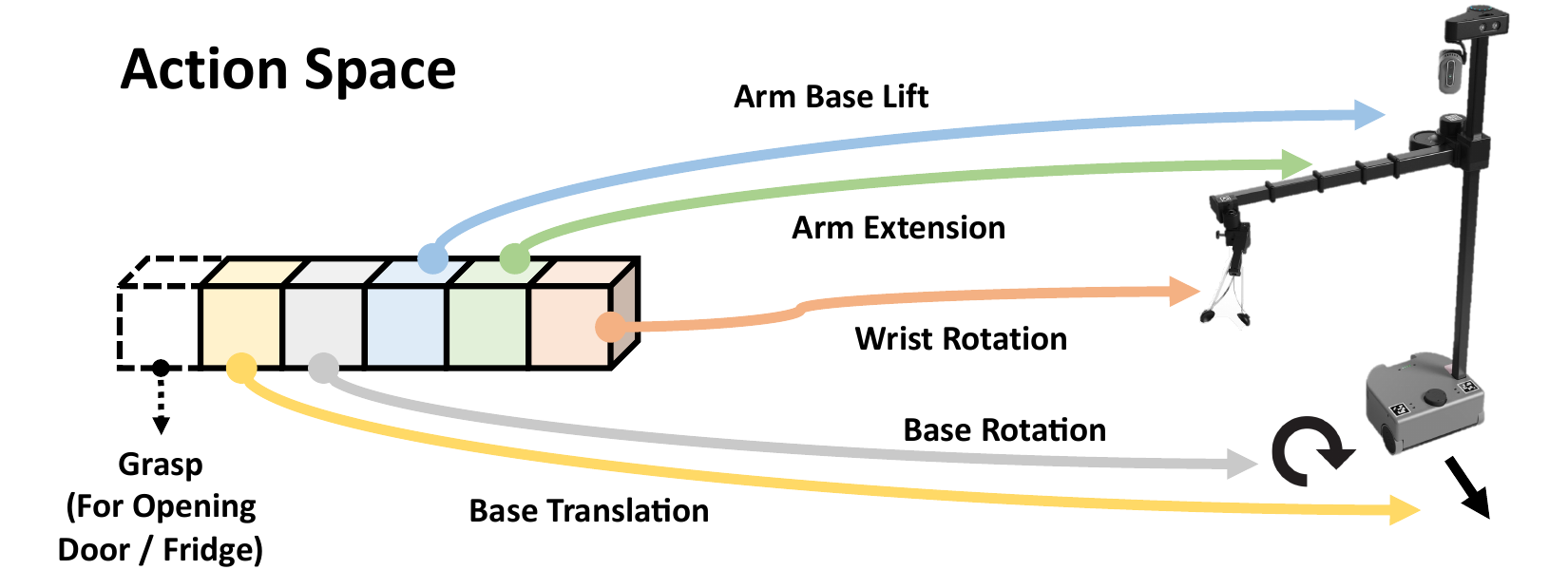}
    \end{center}
    \vspace{-0.2in}
    \caption{\small{Our controller controls all DOF of the robot at every step.
    }}
    \vspace{-0.25in}
    \label{figure:action_space}
\end{figure}

Humans, on a daily basis, perform long-horizon tasks that necessitate close coordination between navigating the body and manipulating objects within the environment. 
While we can effortlessly coordinate actions between our hands and feet, teaching robots to do the same is a very challenging problem. 
To study this problem, first, we introduce a suite of tasks, named the \suite, consisting of atomic tasks that one might want a household robot to perform, as in Fig.~\ref{fig:teaser}. These tasks require an agent to navigate their world while manipulating objects.

\mypara{1) Opening Door.} In most living spaces, navigating through doors is a key task. This involves identifying and operating the door handle by pulling or pushing while adjusting the body position to avoid collision with the door. 
The suite includes both \textit{Opening Door (Push)} and \textit{Opening Door (Pull)}.

\mypara{2) Cleaning Table.} A common chore is cleaning surfaces like tables and counters which requires coordinated movement to clean the entire surface area effectively. The suite features \textit{Cleaning Table} to represent such activities.

\mypara{3) Opening Fridge.} Operating household appliances is crucial for a robotic assistant. This task involves locating and opening an appliance like a fridge, including navigation and manipulation. The suite includes \textit{Opening Fridge} as an example of such tasks.

\noindent\textbf{Training Environments.}
Our training leverages the ProcTHOR\cite{deitke2022procthor} environment, preparing agents for our task suite with an emphasis on bridging the simulation-to-reality (Sim2Real) gap through diverse and photorealistic simulations. For \textit{Opening Door} tasks, we use 700 houses featuring two-room configurations with openable doors. \textit{Cleaning Table} is trained with 2000 houses, varying from one to three rooms, each equipped with a table. For fridge-opening scenarios, 1400 houses with realistically furnished kitchens are employed. Environmental factors like textures and lighting are randomized to enhance generalization capabilities, ensuring effective real-world policy application.

\section{\methodNameLong (\methodName)}
\label{sec:model}

Efficiently coordinating navigation and manipulation in mobile robots presents a significant challenge, particularly when using RL in large action spaces. Prior studies often split tasks into discrete navigation and manipulation stages, or alternately cycle between these modes, limiting the action space at any given timestep (\cite{saycan2022arxiv, brohan2023rt2, yenamandra2023homerobot, xia2021relmogen, sun2021fully, wu2023tidybot}). This formulation of the action space prevents the robot from performing challenging tasks such as those in \suite. Our approach employs the Stretch \cite{kemp2022design}, which navigates and manipulates concurrently, leveraging a 5-dim continuous action space for integrated base and arm movements. As shown in Fig.~\ref{figure:action_space}, our action space includes base movements (translation and rotation), arm height and extension adjustments, and wrist rotation, with an additional dimension for grasp actions in specific tasks like door or fridge opening.

\subsection{Simulated Environment.}
\label{sec:simulated_env}
Our simulation utilizes ProcThor \cite{deitke2022procthor} for its superior visual diversity, addressing a gap in environments like IsaacGym~\cite{makoviychuk2021isaac} and Mujoco~\cite{todorov2012mujoco} that lack extensive visual variation critical for real-world skill transfer. To enable the robot to learn the diverse tasks described in Sec~\ref{sec:tasks}, We augmented the simulation with procedural door/fridge opening logic and photo-realistic table cleaning functionality.

\noindent\textbf{Procedural Door/Fridge Opening:} 
One of our contributions is modifying ProcThor to enable physically realistic articulation manipulation. In the ProcThor\cite{deitke2022procthor} simulation environment, doors and fridges are treated as semantic objects, and their open and closed states are not governed by any physical rules. Instead, we estimate their status at each simulation step based on the robot's actions. For pushing actions, the system calculates the direction the robot's end-effector is moving towards. If this direction aligns with the door's normal vector, the door is opened incrementally in the same direction as the end-effector's movement. In the case of pulling actions, the door's movement is determined after the robot command is executed. The door opens in small segments if the end-effector's movement direction aligns with the door's normal vector, provided the robot has already secured the doorknob with a magnetic grasper. Additionally, collision detection algorithms are in place to prevent the door from colliding with the robot's body during these interactions.

\noindent\textbf{Photo-realistic Table Cleaning:} 
We developed a highly realistic table-cleaning simulation in ProcThor\cite{deitke2022procthor}. This process involves randomly placing a variety of lifelike dirt objects on a designated table at the start of each episode and attaching a sponge to the robot's end-effector at the start of the episode. During each simulation step, any dirt objects that come into contact with the attached sponge are eliminated. The quantity of dirt objects removed is then tallied for the purpose of reward computation in RL training. This method of simulating photorealistic dirt objects not only allows the robot to visually detect dirt on the simulated table but also to recognize the removal of dirt upon cleaning.

\subsection{Robot Perception}
We equip the robot with both RGB visual observations as well as the proprioceptive state of the robot, which is available on most robots and easy to obtain.

\noindent\textbf{Multi-camera RGB Observations:}
In contrast to prior work utilizing a mix of RGB, depth cameras, and LIDAR for mobile manipulation, our approach exclusively uses RGB cameras due to their cost-efficiency, power efficiency, and robustness against environmental noise. RGB cameras have been proven to provide ample scene semantics for navigation and manipulation tasks, supported by advances that have narrowed the simulation-to-reality visual gap (\cite{xia2021relmogen, deep-wbc, saycan2022arxiv, loquercio2023learning, deitke2022procthor, deitke2023phone2proc, khandelwal2022:embodied-clip}).
Our robot is equipped with two RGB cameras: a Navigation Camera (\textit{Nav Cam}) for forward observation and a Manipulation Camera (\textit{Manip Cam}) aimed at the robot's arm, enabling multi-view environmental awareness, as shown in Fig. ~\ref{figure:method}.
Despite the photorealism in simulation~(\cite{deitke2022procthor, ai2thor}), we address the remaining Sim2Real visual discrepancies through augmentation, further detailed in the appendix (Sec~\ref{sec:supp_vis_aug}).

\noindent\textbf{Proprioception:} 
For tasks requiring object manipulation in \suite, our controller utilizes a 5-dimensional proprioception vector, detailing the arm base's lift, arm extension, and wrist rotation, to enhance manipulation capabilities.

\begin{table*}[t]
\caption{\small{
{\bf Task Completion Evaluation (Top):} Our method outperforms baselines with higher Success Rates and higher Progress. {\bf Efficiency Evaluation (Bottom):} Our method outperforms baselines with higher Progress Speed and shorter Episode Length (Eps-Length).} 
}
\vspace{-0.05in}
\label{table:functionality_and_efficiency}
\centering
\resizebox{\textwidth}{!}{
\begin{tabular}{l|cc|cc|cc|cc}
\toprule

 & \multicolumn{2}{c|}{Cleaning Table} & \multicolumn{2}{c|}{Opening Door (Push)} & \multicolumn{2}{c|}{Opening Door (Pull)} & \multicolumn{2}{c}{Opening Fridge} \\
& Success Rate (\%) & Progress (\%) & Success Rate (\%) & Progress (\%) & Success Rate (\%) & Progress (\%) & Success Rate (\%) & Progress (\%) \\
\midrule
Two-Stage & $0.0 \pm 0.0$ & $4.5 \pm 0.7$ & $0.0 \pm 0.0$ & $8.5 \pm 9.2$  & $0.0 \pm 0.0$& $0.85 \pm 0.15$ & $0.0 \pm 0.0$ & $0.4 \pm 1.1$ \\
ReLMoGen\cite{xia2021relmogen} & $25.1 \pm 20.4$ & $69.6 \pm 9.6$ & $51.7 \pm 31.4$ & $74.5 \pm 15.4$ & $40.7 \pm 10.7$ & $62.6 \pm 8.6$ & $0.0 \pm 0.0$ & $6.5 \pm 8.8$ \\
\methodName(Ours) & $\mathbf{36.7 \pm 22.9}$ & $\mathbf{76.0 \pm 11.3}$ & $\mathbf{80.0 \pm 19.9}$ & $\mathbf{89.7 \pm 5.0}$ & $\mathbf{51.2 \pm 12.2}$ & $\mathbf{69.8 \pm 2.7}$ & $\mathbf{20.0 \pm 3.1} $ & $\mathbf{46.2 \pm 5.4}$  \\
\bottomrule
\end{tabular}
}

\vspace{0.05in}

\resizebox{\textwidth}{!}{
\begin{tabular}{l|cc|cc|cc|cc}
\toprule
 & \multicolumn{2}{c|}{Cleaning Table} & \multicolumn{2}{c|}{Opening Door (Push)} & \multicolumn{2}{c|}{Opening Door (Pull)} & \multicolumn{2}{c}{Opening Fridge} \\
& Progress Speed$\uparrow$ & Eps-Length$\downarrow$ & Progress Speed$\uparrow$ & Eps-Length$\downarrow$ & Progress Speed$\uparrow$ & Eps-Length$\downarrow$ & Progress Speed$\uparrow$ & Eps-Length$\downarrow$ \\
\midrule
Two-Stage & $0.05 \pm 0.01$ & $500.0 \pm 0.0$ & $0.09 \pm 0.09$ & $500.0 \pm 0.0$ & $0.00 \pm 0.00$ & $500.0 \pm 0.0$ & $0.0 \pm 0.0 $ & $500 \pm 0$ \\
ReLMoGen\cite{xia2021relmogen} & $1.01 \pm 0.42$ & $443.1 \pm 53.2$ & $1.87 \pm 1.10$ & $352.8 \pm 110.7$ & $1.28 \pm 0.49$ & $401.0 \pm 44.2$ & $0.065 \pm 0.088$ & $500 \pm 0$ \\
\methodName(Ours) & $\mathbf{1.34 \pm 0.56}$ & $\mathbf{410.1 \pm 58.7}$ & $\mathbf{3.98 \pm 2.05}$ & $\mathbf{218.1 \pm 117.4}$ & $\mathbf{1.64 \pm 0.59}$& $\mathbf{370.2 \pm 54.8}$ & $\mathbf{0.847 \pm 0.140}$ & $\mathbf{446.6 \pm 19.8}$ \\
\bottomrule
\end{tabular}
}
\vspace{-0.15in}
\end{table*}

\subsection{\methodName Architecture}
\label{sec:main_network_arch}

As shown in Fig.~\ref{figure:method}, \methodName starts with a DINOv2 visual backbone that processes input from both RGB cameras. These inputs are further refined by two convolutional visual encoders, $E_{Nav}$ and $E_{Manip}$, for navigation and manipulation visual observations, respectively, to generate specific visual embeddings. 
Visual embeddings are concatenated with the robot's proprioception and then fed into a recurrent unit to add memory, followed by a linear layer to produce the robot action. More details can be found in Sec~\ref{sec:supp_network_arch}.
Leveraging a pre-trained visual backbone, as demonstrated by previous research within AI2-THOR~(\cite{khandelwal2022:embodied-clip, deitke2022procthor, deitke2023phone2proc,spoc2023}), offers significant advantages in performance and real-world generalization. This choice is informed by the encoder's proven ability to capture detailed semantic information from visual observations and its success in bridging the simulation-to-reality gap, thanks to its self-supervised training on a diverse web-scale dataset. 

\subsection{Reward Function}
\label{sec:main_reward_function}
We use the following reward function for all tasks:
\vspace{-0.1in}
\begin{equation*}
    R = w_{\text{nav}} * R_{\text{nav}}+  w_{\text{manip}} * R_{\text{manip}} + w_{\text{efficiency}} * R_{\text{efficiency}}
\end{equation*}
where $w_{*}$ are coefficients for different reward terms. 
We provide high-level descriptions of each term below and more details are provided in Sec~\ref{sec:supp_reward_function}.

\noindent\textbf{Navigation Reward $R_{\text{nav}}$:} 
This reward is proportional to the distance the robot moved towards the target object (door/fridge for the \textit{Opening Door/Fridge}, or table for \textit{Cleaning Table}). It provides a reach-target reward when the robot reaches the target and gets cut off afterward.

\noindent\textbf{Manipulation Reward $R_{\text{manip}}$:}
This reward contains two sub-rewards. The first sub-reward encourages the robot to move its end-effector towards the doorknob, the center of the table, and the fridge handle for \textit{Opening Door}, \textit{Cleaning Table}, and \textit{Opening Fridge} tasks respectively. The second sub-reward is proportional to the progress the robot made towards task completion and offers a significant bonus upon successful task completion. Task progress is measured by the degree of openness of the door or fridge for \textit{Opening Door/Fridge} tasks, and the percentage of dirt removed from the table for \textit{Cleaning Table} tasks. 

\noindent\textbf{Efficiency Reward $R_{\text{efficiency}}$:}
This reward incentivizes the robot to complete the task efficiently while maintaining a safe distance from the object and minimizing excessive movement of its end-effector.

\section{Experiment}
\label{sec:experiment}

\begin{figure*}[t]
    \centering 
    \includegraphics[width=0.975\textwidth,clip]{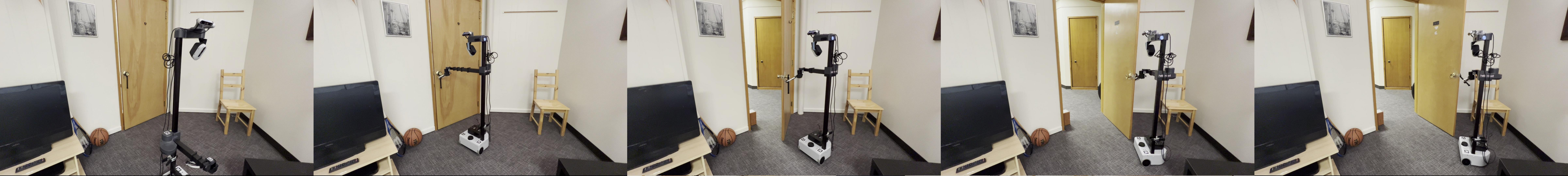}
    \vspace{0.025in}

    \includegraphics[width=0.975\textwidth,clip]{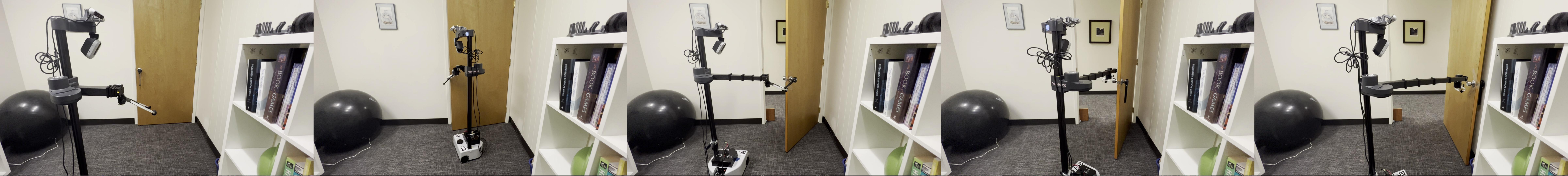}
    \vspace{0.025in}

    \includegraphics[width=0.975\textwidth,clip]{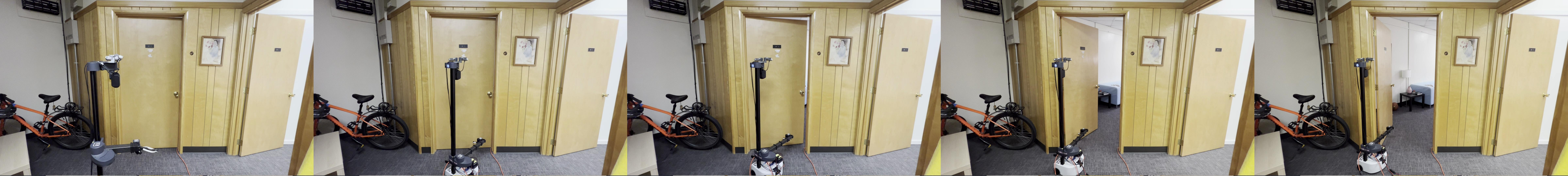}
    \vspace{0.025in}

    \includegraphics[width=0.975\textwidth,clip]{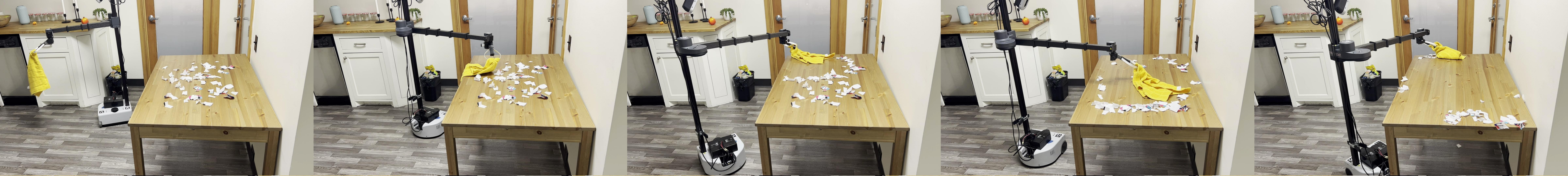}
   
    \vspace{-0.05in}
    \caption{\small{\textbf{Real World:} We deployed the learned controller in a real apartment with a novel layout. Each row shows a single trajectory (from left to right) corresponding to Opening Door (Pull), Opening Door (Pull), Opening Door (Push), and Cleaning Table, respectively. }}
    \label{fig:real_trajs}
\vspace{-0.1in}
\end{figure*}

We evaluated our method on \suite both in simulation and in the real world. 
We deployed the policy (learned in simulation) in a real apartment with multiple rooms \textit{without any adaptation or fine-tuning}.

\subsection{Training and Evaluation Details}
We trained the policies for all methods for each task using DD-PPO(~\cite{wijmans2020ddppo}) for $10^{7}$ steps using AllenAct\cite{AllenAct}. Each experiment was repeated with $3$ different seeds.
We evaluated the policies on $1200$ episodes ($300$ per task). Our evaluations were conducted in 350 unseen houses with novel layouts.

\mypara{Robot Initialization Position in Scene:}
Recognizing the impact of the initial position of the robot on mobile manipulation, we optimized the robot's start points to improve exploration efficiency and task performance. 
If the robot is too close to the object, it struggles to adjust its pose, while, initializing the robot too far from the object negatively impacts its ability to learn required skills.
We place the robot $1-4$ meters away from the table for \textit{Cleaning Table}, and $1-3$ meters away from the door/fridge for \textit{Opening Door/Fridge} tasks.

\mypara{Metrics:} We evaluate the performance of all methods with:

\noindent\textbf{Progress}: This metric (as defined in \ref{sec:main_reward_function}) indicates how much progress the agent made toward completing the task. 
Our tasks are complex and long-horizon, making this an insightful metric, especially where the agent does not reach a high completion rate but makes meaningful progress.

\noindent\textbf{Success Rate}: 
Success was defined by the task progress: over $90\%$ for door opening, $70\%$ for fridge opening (since normally fridges are located in a narrower space and humans don't fully open the fridge), and $75\%$ for table cleaning.

\noindent \textbf{Episode Length}:
Reflects the average steps required for task completion, with episode length capped at $500$ steps.

\noindent \textbf{Progress Speed}: This metric reflects the efficiency of the agent in making progress toward task completion. It is calculated by:
$
\frac{\text{Task Progress}}{\text{Episode Length} / \text{Max Episode Length}}
$

\begin{table*}[t]
\caption{\small{{\bf Proprioception Ablation:} We ablate the necessity of proprioception information on Opening Door tasks. Proprioception information is necessary for tasks requiring precise placement of the end-effector and helpful for other tasks.}}
\vspace{-0.1in}
\label{table:ablation}
\centering
\resizebox{\textwidth}{!}{
\begin{tabular}{l|cccc|cccc}
\toprule
 & \multicolumn{4}{c|}{Opening Door (Push)} & \multicolumn{4}{c}{Opening Door (Pull)} \\
& Success Rate (\%) & Progress (\%) & Progress Speed$\uparrow$ & Eps-Length$\downarrow$ & Success Rate (\%) & Progress (\%) &  Progress Speed$\uparrow$ & Eps-Length$\downarrow$ \\
\midrule
\methodName w/o proprio & $74.6 \pm 1.1$ & $ 83.9 \pm 1.4$ & $2.51\pm 0.17$ & $263.4\pm 15.0$ & $5.2 \pm 9.0$ & $ 11.0 \pm 18.9 $ & $0.16 \pm 0.28$ & $489.5 \pm 18.1$ \\

\methodName & $\mathbf{80.0 \pm 19.9}$ & $\mathbf{89.7 \pm 5.0}$ &  $\mathbf{3.98 \pm 2.05}$ & $\mathbf{218.1 \pm 117.4}$ & $\mathbf{51.2 \pm 12.2}$ & $\mathbf{69.8 \pm 2.7}$ & $\mathbf{1.64 \pm 0.59}$& $\mathbf{370.2 \pm 54.8}$ \\
\bottomrule

\end{tabular}
}
\vspace{-0.1in}
\end{table*}

\subsection{Performance of \methodName in Simulation}

Our experiment results demonstrate that \methodName achieves an average success rate of 47\% across all tasks (As in Table~\ref{table:functionality_and_efficiency}). Specifically, our model attains a high success rate of 80\% for the task of \textit{Opening Door (Push)} in unseen houses, and achieves
average progress rates of 76\%, 70\%, and 46\% for \textit{Cleaning Table}, \textit{Opening Door (Pull)}, and \textit{Opening Fridge}, respectively, indicating significant progress in each task. This achievement is especially impressive given the complex, long-horizon nature of these tasks. 

The success rates across tasks vary due to how the robot interacts with its environment. \textit{Opening Door (Push)} has a high success rate as it’s less affected by environmental factors: the robot can use the space created by the opened door to continue the task. However, \textit{Opening Door (Pull)} and \textit{Opening Fridge} are more challenging. These tasks reduce the available space as they progress, especially in narrow areas, leading to lower success rates despite significant progress. \textit{Cleaning Table} involves complex navigation due to obstacles and furniture around the table, requiring intricate movement patterns. This complexity accounts for its lower performance compared to door-related tasks. For \textit{Opening Fridge}, the task’s difficulty arises from the smaller size of fridge doors and the need for precise end-effector placement. The often confined placement of fridges in rooms adds to the challenge, affecting the robot’s ability to navigate and manipulate effectively.

\begin{figure}[t]
    \centering
    \begin{center}
        \includegraphics[width=\linewidth]{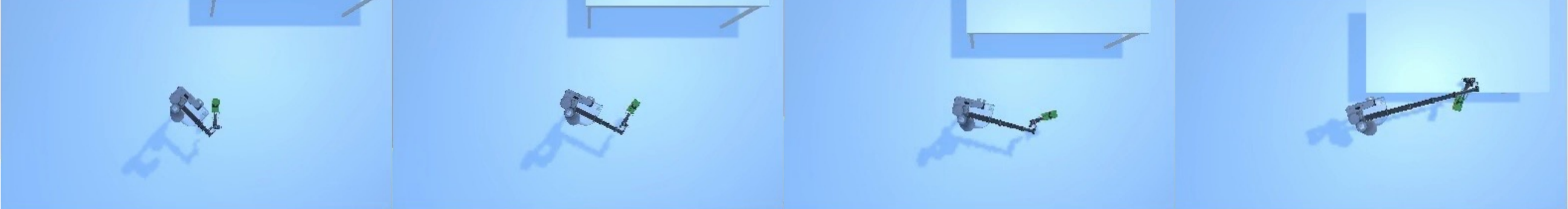}
    \end{center}
    \vspace{-0.15in}

    \caption{\small{\textbf{Cleaning Table in Simulation}
    }}
    \vspace{-0.35in}
    \label{fig:sim_cleaning_table_traj}
\end{figure}

\noindent\textbf{Comparison with the previous approaches.}
We hypothesize that jointly learning and executing navigation and manipulation for mobile manipulation tasks achieves better performance than considering them disjointly. 
To validate this,
we compared our method with two common approaches widely used by the community. For a fair comparison, we use the same network architecture stated in Sec~\ref{sec:model} and the same hyperparameters for baselines.

One common practice, which we note as \textit{Two-Stage} baseline, in embodied AI for mobile manipulation~(\cite{yenamandra2023homerobot, gu2023multiskill, szot2022habitat}) is to decompose the tasks into separate stages of manipulation and navigation. In this work, our \textit{Two-Stage} baseline uses a roughly similar high-level structure as OVMM\cite{yenamandra2023homerobot} in which the robot navigates to the target object and, after reaching the target, switches to the manipulation. While this approach works for a variety of pick-and-place tasks, it fails for more complex tasks, such as those presented in our benchmark. Table~\ref{table:functionality_and_efficiency} shows the very low success rate and progress of the \textit{Two-Stage} baseline. 

Another common approach is executing manipulation and navigation alternatively. At every step controller invokes separately obtained manipulation or navigation skills. 
We adapted ReLMoGen~\cite{xia2021relmogen} to support the embodiment of the Stretch robot. This baseline outputs the target end-effector pose, target base pose in the current robot frame, and choice over navigation or manipulation at every step. Although this approach achieves much higher results than the Two-Stage baseline, there is still a significant 17.6\% (37\% relative) drop in success rate compared to \methodName. 

\noindent\textbf{\methodName is Efficient.}
Confirming our hypothesis, \methodName not only boosts performance but also enhances efficiency. In table cleaning, our method produces efficient motion of extending arms while approaching the table, as shown in Fig. \ref{fig:sim_cleaning_table_traj}. 
Table~\ref{table:functionality_and_efficiency} shows that \methodName makes 32.2\% more progress towards completing the task at each step compared to the baselines on \textit{Cleaning Table}, 113.4\% on \textit{Opening Door (Push)}, and 27.6\% on \textit{Opening Door (Pull)}.

\subsection{\methodName transfers to the real world.}

\begin{table}[t]
\small
\caption{\small{{\bf Real World Evaluation}
}}
\vspace{-0.05in}
\label{table:real_world}
\centering
\begin{tabular}{l|c|c}
\toprule
Task & Success Rate & \# of trials\\
\midrule
Cleaning Table & $66.6\%$    & 12 \\
Opening Door (Pull) & $60.0\%$ & 15 \\
Opening Door (Push) & $70.0\%$ & 10 \\
\bottomrule
\end{tabular}
\vspace{-0.1in}
\end{table}

\begin{table}[t]
\centering
\caption{\small{\bf Ablation Study on Opening Door (Pull)}}
\vspace{-0.05in}
\label{table:additional_ablation}
\begin{tabular}{c|c|c|c}
\toprule
\textbf{Opening Door (Pull)} & Success Rate &  Progress & Reach Door \\
\midrule
No Pretrained DINOv2 & $0\%$  & $0\%$ & $66.3\%$ \\

\midrule
Nav Cam Only & $27.3\%$ & $47.9\%$ & $93\%$ \\
Manip Cam Only  & $0\%$ & $0.235\%$ & $95\%$\\
\midrule
Adapted Skill Trans &  $0\%$ & $0.\%$ & $29.7\%$  \\

\bottomrule
\end{tabular}
\vspace{-0.2in}
\end{table}

Our learned policies were evaluated in a real apartment for \textit{Opening Door (Pull)}, \textit{Opening Door (Push)}, and \textit{Cleaning Table} tasks, showing promising outcomes as in Table~\ref{table:real_world} and Fig. ~\ref{fig:real_trajs}. The Opening Fridge task is not evaluated since the magnetic seal of the fridge is too strong for our robot to pull. To compensate for the simulation's simplified magnetic gripper grasping, a heuristic function for doorknob grasping was implemented in the real world, effectively bridging the simulation-to-reality gap. When our controller initiates a grasp action, the heuristic function is invoked to grasp the doorknob. 

Our controller successfully pulled the door fully open in 9 out of 15 attempts in three different rooms. Interestingly, we observed that \methodName exhibited two distinct styles of pulling the door to adapt to the layout differences in each room. When there was enough open space around the door, it continued pulling until the door was fully opened, as shown in the 1st row in Fig. \ref{fig:real_trajs}. However, when the door was located next to a wall, the agent first pulled the door to start the opening process and then switched to pushing to fully open it (2nd row, Fig. \ref{fig:real_trajs}). These emergent behaviors showcase the spatial reasoning and scene-understanding abilities of our controller. Similarly, our controller successfully pushed doors fully open 7 times in 4 different rooms out of 10 trials.

For the cleaning table task, we placed the table in four different locations across two rooms, with evenly distributed paper pieces on it for the robot to clean. Our robot navigated to the table and continuously moved its end-effector to clean the entire table, as shown in Fig. ~\ref{fig:real_trajs}. Out of 12 trials, our controller successfully cleaned $66.6\%$ of the table's surface area in eight instances.

\begin{figure}[t]
    \begin{center}
        \includegraphics[width=0.35\textwidth]{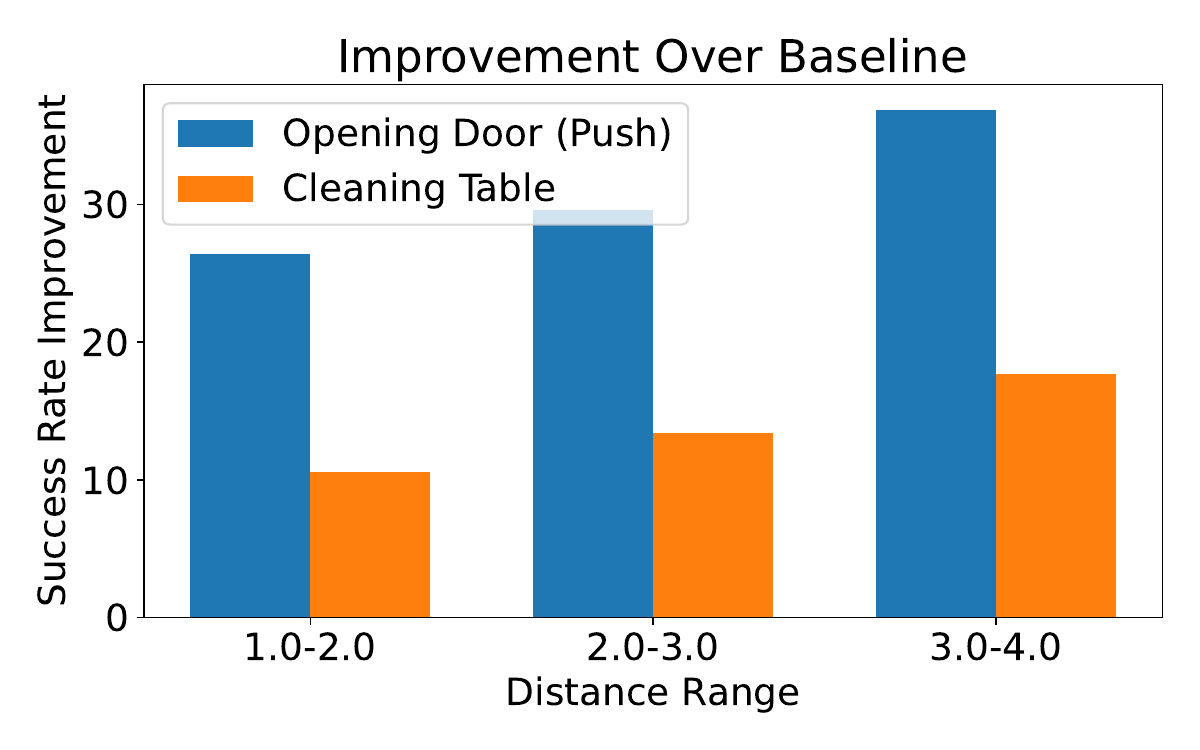}
    \end{center}
    \vspace{-0.25in}
    \caption{\small{\textbf{Improvement over Range:} 
    The improvement \methodName obtained over ReLMoGen grows as the distance increases. 
    }}
    \vspace{-0.2in}
    \label{fig:success_rate_over_range}
\end{figure}

\subsection{Ablation Study of \methodName}

\noindent\textbf{More efficient in longer horizon tasks.}
we analyzed the success rate of both \methodName and ReLMoGen\cite{xia2021relmogen} in the \textit{Cleaning Table} and \textit{Opening Door (Push)} tasks across varying initial distances from the target object of interest (aka. table and door), ranging from $1 \sim 4 \meter$. As in Fig. ~\ref{fig:success_rate_over_range}, we observed that the improvement of \methodName over the baseline becomes larger as the initial distance to the target object increases. This trend underscores \methodName's enhanced efficiency, particularly in tasks with longer horizons, where the coordinated integration of navigation and manipulation is pivotal in successful task completion.

\noindent\textbf{Proprioception.}
Our comparison of \methodName against a proprioception-less variant in door-opening tasks illustrates the proprioception significance (in Table~\ref{table:ablation}). We also observe $-33.7\%$ / $-20\%$ success rate drop, and $-8.3\%$ / $-26.2\%$ progress drop for \textit{Cleaning Table} and \textit{Opening Fridge} without proprioception. The absence of proprioceptive feedback notably diminishes performance, especially in \textit{Opening Door (Pull)}, where precise end-effector positioning is necessary for pre-grasping the knob. This effect is less pronounced but still present in \textit{Opening Door (Push)}, emphasizing proprioception's overall contribution to task performance and efficiency, even when precise manipulation is not as critical.

\noindent\textbf{Multi-View Visual Observation.} We had initial experiments with a single-camera setup showing a significant performance drop compared with multi-camera ones (provided in Table~\ref{table:additional_ablation}). This indicates that complex tasks require simultaneous visual input for both navigation and manipulation, validating our approach of integrated navigation and manipulation. 

\noindent\textbf{Pretrained Visual Encoder} We substituted the DINOv2 encoder with a trainable CNN. This modification led to a significant drop in success rate and progress (Tab.\ref{table:additional_ablation}), underscoring the critical role of a pretrained visual encoder

\noindent\textbf{Transformer Variant.} We adapted Skill Transformer\cite{huang2023skill} and evaluated it on the most complicated task \textit{Open Door (Pull)} (As in Table ~\ref{table:additional_ablation}). It performs poorly, likely due to the low-sample efficiency of training a large transformer with a ResNet18 encoder from scratch (following the original paper), which is unsuitable for our online RL setting. 
The original paper trains different low-level skills separately and uses a skill transformer for coordinating skills, which is significantly different from our task setting.

\section{Conclusion}
\label{sec:discussion}

In this work, we developed a simulation for complex mobile manipulation tasks, ranging from opening doors/fridges to cleaning tables in daily scenes, using ProcThor\cite{deitke2022procthor}. Our \methodName enables robots to solve these tasks using only RGB visual observation and proprioception of the robot through end-to-end learning. Our pipeline significantly outperforms previous baselines in simulation and has successfully transferred to real-world apartments with novel layouts, without any fine-tuning. Code for our benchmark and our method will be released.

\noindent\textbf{Limitation:} 
Despite its success, our \methodName faces limitations related to the kinematic and physical capabilities of the robot, restricting its ability to handle more dynamic real-world tasks, such as lifting heavy objects.

{\noindent \textbf{Acknowledgements.} 
We would like to thank the Thor Team: Winson Han, Eli VanderBilt, and Alvaro Herrasti, for their support in setting up the simulations. Special thanks to Winson Han for his expertise and assistance in the visualization of this work.
}\label{sec:ack}
\bibliographystyle{IEEEtran}
\bibliography{main}

\maketitlesupplementary

\section{Visual Augmentation}
\label{sec:supp_vis_aug}

Our applied augmentations include:
\textbf{ColorJitter}: Adjusts brightness (0.4), contrast (0.4), saturation (0.2), and hue (0.05); \textbf{GaussianBlur}: Applies a blur effect with a kernel size of (5,9) and sigma range of (0.1,2); \textbf{RandomResizedCrop}: Resizes the input with a scale range of (0.9,1); \textbf{RandomPosterize}: Reduces color depth, applied with varying bits (7,6,5,4) and a probability of 0.2 for each setting; \textbf{RandomAdjustSharpness}: Enhances or reduces the sharpness with a factor of 2, applied with a probability of 0.5.

\section{Network Architecture}
\label{sec:supp_network_arch}
Our RGB observations initially get processed by DinoV2 model, generating $768 \times 16 \times 16$ features which pooled to a reduced dimension of $768  \times 7 \times 7$. The processed features are passed through separate visual encoders. This results in two distinct sets of $16 \times 7 \times 7 $ latent visual features corresponding to each camera view. 
The key hyperparameters are as follows:
\begin{center}
\small{
\begin{tabular}{l|c}
     Hyperparameter  & Value \\
     \hline
    RGB Input & $384\times224$ \\
    Pretrained Visual Encoder & dinov2\_vits14 \\
    $E_{Nav}$  & 2 Conv(1x1, 16 channels) \\
    $E_{Manip}$ & 2 Conv(1x1, 16 channels) \\
    GRU & 1 GRU layer with 512 units \\
    Policy($\pi$) & 1 FC with 512 units \\
    Value($v$) & 1 FC with 512 units \\
    Proprioception dim & 5 \\
    latent feature dim & $7*7*16 * 2 + 5 = 1573$
\end{tabular}
}
\end{center}

\section{RL training Hyperparameters}
\label{sec:supp_hyper_parameter}
\begin{center}
\small{
\begin{tabular}{l|c}
     Hyperparameter & Value \\
     \hline
     Non-linearity & ReLU \\
     Policy initialization & Standard Gaussian\\
     \# of samples per iteration & 640 \\
     Discount factor & .99 \\
     Parallel Environment & 20 \\
     Batch size & 640 \\
     Optimization epochs & 4 \\
     Clip parameter & 0.1 \\
     Policy network learning rate & 5e-5 \\
     Value network learning rate & 5e-5 \\
     Entropy & 0.0025 \\
\end{tabular}
}
\end{center}

\section{Reward Function}
\label{sec:supp_reward_function}
In this section, we provide the formulation of our reward and hyperparameters for different tasks in Sec~\ref{sec:main_reward_function}.

\noindent\textbf{Navigation Reward $R_{\text{nav}}$:} 
\begin{equation*}
    R_{\text{nav}} = \delta w_{\text{nav shaping}} max(d_{\text{closest}} - d_{\text{current}}, 0) + \gamma  R_{\text{reach target}} 
\end{equation*}
The variable  $\delta$ is $1$ if the robot hasn't reached the target position and $0$ otherwise. Similarly, $\gamma$ is 1 if the robot reaches the target at the current step, and 0 otherwise.
$d_{\text{closest}}$ is the closest distance robot reached before, $d_{\text{current}}$ is the current distance between robot and target object. For all tasks, we use $w_{\text{nav shaping}}=2, R_{\text{reach target}}=2, w_{\text{nav}}=1$ .

\noindent\textbf{Manipulation Reward $R_{\text{manip}}$:}
\begin{align*}
R_{\text{manip}} = &\delta w_{\text{manip shaping}}  R_{\text{manip shaping}}\\
 + & w_{\text{progress}} \Delta P + \gamma R_{\text{finish task}}
\end{align*}
\begin{align*}
    R_{\text{manip shaping}} = &exp(-5\times d_{\text{ee current})} \\
    &\times 1000 \times max(d_{\text{ee cloest}} - d_{\text{ee current}}, 0))
\end{align*}
The variable  $\delta$ is $1$ if the end effector has not reached the target region, and $0$ otherwise. Similarly, $\gamma$ is 1 if the robot completes the task at the current step, and 0 otherwise.

$d_{\text{closest}}$ is the closest distance the end-effector reached before towards the specified region of the target object, $d_{\text{current}}$ is the current distance between the end-effector and the specified region of the target object, $P$ is the task progress. For all tasks, we use $w_{\text{manip}}=1, w_{\text{manip shaping}}=0.02, R_{\text{finish task}}=20$. For opening door/fridge tasks, we use $w_{\text{progress}} = 80$. For the cleaning table task, we use $w_{\text{progress}=100}$. For opening the door (pull) and opening the fridge task, we provide an additional $w_{grasp} = 2$ for grasping the door knob or the edge of the fridge. We consider the grasp successful if the agent has issued the grasp action and the distance between the end-effector and object is smaller than $0.2 \meter$.

\noindent\textbf{Efficiency Reward $R_{\text{efficiency}}$:}
\begin{align*}
R_{\text{efficiency}} = & R_{\text{step penalty}}\\
 + & w_{\text{ee moved}} ||d_{\text{ee moved}}|| + \gamma R_{\text{invalid action}}
\end{align*}
where $d_\text{ee moved}$ is the distance end-effector moved at the current step, and 
$\gamma$ is 1 if the current action failed, and 0 otherwise.
Current action fails if the robot collides with semantic objects in the apartment in simulation, such as a wall or table. For all tasks, we use $R_{\text{step penalty}} = -0.01, w_{\text{ee moved}} = -0.01, R_{\text{invalid action}} = -0.01, w_{\text{efficiency}} = 1$

\noindent\textbf{Hardware Platform:} Our simulation and real-world experiment are conducted with Stretch Robot~\cite{kemp2022design} and we use RealSense D455 for our RGB observation.

\mypara{Future Works:} 
In future work, we aim to enhance the capabilities of \methodName by integrating more complex and dynamic tasks into our task suite. We aim to explore the potential of \methodName in environments with even longer task horizons and more varied challenges.

\end{document}